%% file: main.tex
\documentclass{article}
\usepackage[T1]{fontenc}
\usepackage{iclr2026_conference,times}
\usepackage{amsmath,amssymb,booktabs,graphicx,microtype}
\usepackage[colorlinks=true,linkcolor=blue,citecolor=blue,urlcolor=blue]{hyperref}
\iclrfinalcopy
\title{\raggedright Delayed Supervision for\\Test-Time Language Models}
\author{Jinha Kim\thanks{Equal contribution. Work done at MIT.}\quad Taksh Kothari\footnotemark[1]\\MIT CSAIL}

\newcommand{\ntp}{\mathcal{L}_{\mathrm{NTP}}}
\newcommand{\qa}{\mathcal{L}_{\mathrm{QA}}}
\begin{document}
\maketitle
\lhead{Preprint}
\begingroup\renewcommand{\thefootnote}{}\footnotetext{\textcopyright\ 2026 Jinha Kim. All rights reserved, subject to applicable law and granted licenses. We welcome research collaborations; please contact the authors. Please cite this preprint when using or building upon its methods, experimental designs, or results.}\endgroup
\begin{abstract}
Test-time language models adapt a compact memory while processing the input sequence. This perspective encompasses nonlinear fast-weight learning in LaCT, associative delta-rule updates in DeltaNet, and generalized delta-rule state updates in RWKV-7. Training these models to predict the next token does not explicitly require a fact to remain accessible after many subsequent memory updates. We study delayed supervision for this test-time memory: during post-training, ask a simulator-grounded question only after a long interval of unrelated events, and supervise its answer alongside ordinary next-token prediction. Questions are evaluated on disposable branches, so their answers never enter the continuing event stream. The construction distinguishes retention from revision: a retained fact must remain valid throughout the delay, whereas a revised fact must be answered with its latest value. We evaluate this approach on LaCT-760M and plain DeltaNet-1.3B using TextWorld training trajectories and shared BABILong and RULER evaluation panels, and include a separately reported RWKV-7 comparison. Relative to event-only training, delayed QA improves BABILong by 5.48 percentage points for LaCT and 1.32 points for DeltaNet, and single-needle RULER by 1.45 and 3.27 points, respectively. The RWKV-7 comparison reports gains of 4.60 and 7.00 points on its own panels. These results support delayed semantic supervision as a practical outer training objective for usable test-time memory, while leaving open how much of the benefit derives specifically from delay rather than general question-answering and answer-termination supervision.
\end{abstract}

\section{Introduction}
A test-time language model continually adapts its memory to the sequence it is processing. Over a long interaction, that memory must do more than preserve text. It must retain facts that remain relevant, update facts that change, and recover the current state when a later question demands it. Consider an agent that observes a key being placed in a pantry, then moved to a bedroom. After thousands of tokens about other objects, a useful memory must answer ``bedroom.'' Remembering the first location indefinitely is as incorrect as forgetting the key altogether.

The connection between sequence modeling and test-time learning provides a common way to study this problem. Linear attention admits efficient recurrent computation \citep{linear} and a fast-weight interpretation \citep{fastweights}. Test-time training (TTT) makes the state an adapted predictor whose parameters change with the observed context \citep{ttt}. LaCT implements nonlinear fast-weight adaptation; DeltaNet implements error-correcting associative updates; and RWKV-7 extends the delta rule with richer state evolution \citep{lact,delta,rwkv}. We refer to these as \emph{test-time language models} through their shared use of context-adapted memory, while retaining their distinct update mechanisms. Their efficiency makes long streams feasible, but their memory must continually decide which information to preserve and which to overwrite.

Most language-model training supplies this decision indirectly through next-token prediction (NTP). This objective can reward long-range dependence, but it need not provide a strong signal for an old fact when that fact is not needed to predict subsequent local text. A stream can contain many predictable transitions without ever asking the model to retrieve the fact that matters later. The resulting gap is between \emph{continuing the observed stream} and \emph{answering a delayed question about its state}.

We address this gap by supervising what test-time memory must make available later. Our delayed question-answer objective supplies that supervision through the existing language-model head. From a generated environment, we select a question with a verifiable answer, allow the event stream to continue, and ask the question only after substantial intervening activity. The answer loss is added to event NTP. Its gradient can influence the earlier writes, intervening state transitions, and eventual readout that together determine the delayed answer. The native test-time update rule is unchanged: delayed supervision trains the slow parameters that shape writing, retention, and reading. No new memory module is required, and evaluation uses ordinary language generation.

The contribution is an objective and a controlled construction for training it. First, we define delayed semantic probes whose validity is checked throughout the interval, including explicit handling of revisions. Second, we combine answer supervision with an unchanged event objective and isolate probes from the continuing trajectory. Third, we measure transfer from TextWorld to long-context QA and retrieval using original pretrained and event-NTP baselines. Our focus is the practical value of the combined objective. The present comparisons do not identify delay as the sole cause of improvement.

\begin{figure}[t]
\centering\includegraphics[width=\linewidth]{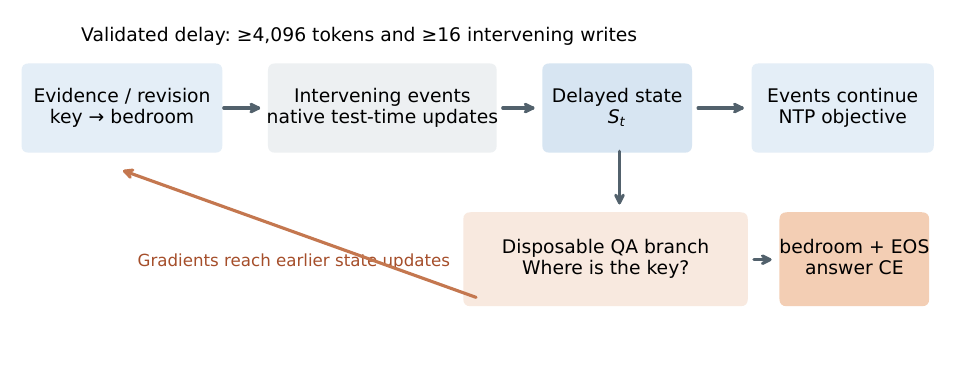}
\caption{Delayed semantic supervision. The event stream updates the model normally. A disposable branch asks a question about an earlier, still-valid fact or the latest revision. Only the branch's answer and EOS receive auxiliary cross-entropy targets. The main trajectory continues without seeing that answer. The displayed delay threshold describes the LaCT/DeltaNet construction; the RWKV-7 report uses a longer minimum interval.}
\label{fig:method}
\end{figure}

\section{Test-time language models}
\label{sec:background}
We use \emph{test-time memory} to mean sequence-specific fast weights or associative state updated from the current input during inference. This is distinct from updating the pretrained slow parameters using test labels. Let $x_{1:T}$ denote a token stream and $S_t$ the model's memory state after token $t$. The models considered here implement
\begin{equation}
 S_t=U_\theta(S_{t-1},x_t),\qquad
 p_\theta(x_{t+1}\mid x_{\leq t})=R_\theta(S_t,x_{\leq t}^{\mathrm{local}}),
\end{equation}
where $\theta$ denotes learned parameters and the optional local context accounts for architectures that combine recurrence with windowed attention. Here $S_t$ evolves at test time, while $\theta$ remains fixed during benchmark inference. The update may run token by token or over native chunks.

In TTT layers, part of $S_t$ is itself a predictor's parameter vector. Context-dependent training updates this fast predictor during sequence processing \citep{ttt}. A schematic gradient update is
\begin{equation}
 W_{t+1}=W_t-\eta_t\nabla_W\ell_{\mathrm{inner}}(W_t;x_t).
\end{equation}
The fast weights are sequence-specific memory; the slow parameters $\theta$ learn how to construct the update and use its output. The inner update can operate over chunks rather than single tokens, and need not be plain gradient descent in every architecture. LaCT makes large-chunk adaptation practical for nonlinear fast-weight networks \citep{lact}; Titans further explores learned memory with explicit retention mechanisms \citep{titans}.

The fast-weight view also connects delta-rule architectures to online learning of an associative predictor. For example, in a column-vector convention, delta-rule memory minimizes a local associative error,
\begin{equation}
 \ell_t(W)=\tfrac12\|Wk_t-v_t\|_2^2,\qquad
 W_t=W_{t-1}+\eta_t(v_t-W_{t-1}k_t)k_t^\top.
\end{equation}
The prediction error corrects the previous key--value mapping rather than merely accumulating an outer product \citep{fastweights,delta}. RWKV-7 generalizes this family with vector-valued gating and in-context learning rates, and distinct removal and insertion keys \citep{rwkv}. Its native update is not assumed to equal the simple squared-error step above. Across these parameterizations, the common question is whether supervising the later semantic use of test-time memory can improve how it learns from context.

Our method operates on the outer objective that trains these test-time learners. We preserve each backbone's native state update, including LaCT's fast-weight adaptation and the delta-rule updates of DeltaNet and RWKV-7. Auxiliary questions are supplied during post-training, where their gradients teach the slow parameters how to form a useful adaptive memory. At benchmark inference, the memory continues to adapt to the observed sequence without ground-truth probe labels or any additional auxiliary optimizer.

\section{Delayed supervision of test-time memory}
\label{sec:method}
\subsection{A state-grounded task}
An environment trajectory consists of events $e_{1:M}$ and simulator states $z_{0:M}$. A question $q$ defines an answer function $a(q,z)$ over those states. We serialize observable events into the model's input stream and attach a probe $(s,t,q,y)$, where $s$ marks sufficient evidence, $t>s$ is the delayed endpoint, and $y=a(q,z_t)$ is the target answer.

\paragraph{Retention and revision.}
For a retention probe, the answer and its relevant dependencies must remain unchanged throughout the interval. For a revision probe, an earlier value must genuinely differ from the current target, and the delay begins after the last relevant revision. Both cases require the current target to remain valid until the question is asked. This prevents a superficially long interval from concealing a recent refresh of the queried fact or an obsolete answer.

\paragraph{Two complementary measures of delay.}
We record both the number of intervening native tokens, $D_{\mathrm{tok}}$, and the number of successful state-changing events, $D_{\mathrm{write}}$. These quantities capture different aspects of interference: a verbose event can add many tokens without adding many state mutations. In the LaCT/DeltaNet construction, accepted pairs require at least 4,096 native tokens and 16 intervening writes. Counts near 16, 32, 64, and 128 writes are sampling strata, not exact acceptance conditions. A valid 27-write example is assigned to the nearest target bucket rather than discarded.

\paragraph{Question families.}
The TextWorld environment provides structured ground truth for questions about locations, inventories, attributes, counts, and relations composed through holders or containers \citep{textworld}. These are questions about the training environment, not answers imported from the external benchmarks. The core DeltaNet set contains eight paired questions per world, divided equally between retention and revision. The LaCT run also includes four terminal questions, as detailed in Section~\ref{sec:setup}.

\subsection{An auxiliary answer objective}
Let $\mathcal{P}(e)$ be the probes attached to one event trajectory. Ordinary event NTP is
\begin{equation}
 \ntp(e)=-\frac{1}{T-1}\sum_{i=1}^{T-1}
 \log p_\theta(x_{i+1}\mid x_{\leq i}).
\end{equation}
For probe $j$, append its question to the event-only prefix ending at $t_j$. Let $\bar y_j$ be the answer tokens followed by EOS. The probe loss is
\begin{equation}
 \ell_j=-\frac{1}{|\bar y_j|}\sum_{r=1}^{|\bar y_j|}
 \log p_\theta(\bar y_{j,r}\mid x_{\leq t_j},q_j,\bar y_{j,<r}),\qquad
 \qa(e)=\frac{1}{|\mathcal{P}(e)|}\sum_j\ell_j.
\end{equation}
The combined objective averages over worlds:
\begin{equation}
 \boxed{\mathcal{L}(\theta)=\mathbb{E}_{e}
 \left[\alpha\ntp(e)+\beta\qa(e)\right].}
 \label{eq:loss}
\end{equation}
Context and question tokens receive no auxiliary targets. They remain in the differentiable computation, so the answer loss can train how evidence is encoded and survives subsequent updates. Separate token normalization for event and answer losses prevents the number of context tokens from mechanically diluting each short answer. It does not guarantee equal gradient magnitude; the auxiliary weight remains an important model-specific choice.

\subsection{Disposable branches and ordinary inference}
Each probe starts independently from an event-only prefix. Its question, teacher-forced answer, and resulting branch state are discarded after scoring. The next probe never receives an earlier probe's answer, and the main event trajectory receives its event loss once per world. Conceptually this is a fork of the causal computation; an implementation may replay the prefix to retain the native model's exact computation.

The supervisory target is the future usefulness of the adapted memory, expressed through a semantic answer. The method is not delayed reconstruction of an earlier sentence. Its target is the \emph{current semantic answer}, which may differ from the wording of the evidence and must reflect intervening revisions. It is also not a hidden-state classifier: the auxiliary loss uses the model's existing language-model head. At inference, we remove the training-only branch construction and use normal generation. Consequently, any external improvement must transfer into the backbone's ordinary answer pathway.

\section{Experimental setup}
\label{sec:setup}
\subsection{Models and controls}
We compare three conditions: \textbf{Original}, the untouched public checkpoint; \textbf{NTP}, post-training on event text alone; and \textbf{Delayed QA}, post-training on event text plus Equation~\ref{eq:loss}. Every trained arm starts from its comparison's public initialization. We study LaCT-760M and plain DeltaNet-1.3B, and report an additional RWKV-7-1.5B comparison from the authors' corrective-run record. The RWKV-7 original-model score is unavailable for that panel and is not inferred.

LaCT and DeltaNet use the same corpus of 512 TextWorld worlds, two passes through the corpus, an effective batch of eight worlds, and 128 outer optimizer updates. Native state updates continue at the architecture's own frequency; an outer optimizer update does not occur after every question. LaCT uses eight delayed paired probes plus four terminal probes per world. DeltaNet uses eight paired probes, with delayed endpoints extended to the latest valid point. Its realized evidence-to-question gaps range from 4,105 to 20,354 native tokens, with a median of 10,099; 92.3\% exceed 8,192 tokens.

The LaCT QA objective uses $(\alpha,\beta)=(0.5,0.5)$. DeltaNet uses $(1,0.0158474)$, fixed from training-only gradient calibration. Both use full-model AdamW with peak learning rate $10^{-5}$, ten warmup updates, and cosine decay to $10^{-6}$. The RWKV-7 comparison uses $(1,0.01)$, batch four, learning rate $10^{-5}$, and 16 updates over a matched sequence of 64 world exposures. It uses two probes per world at approximately 14--15K-token endpoints, with a minimum 8,192-token immediate-to-delayed gap. These settings are matched within the reported RWKV comparison, not to the LaCT/DeltaNet exposure budget.

The three architectures therefore provide separate within-backbone tests rather than a fully controlled architecture ranking. Pretraining, tokenization, probe budgets, delay placement, and loss weights differ across them. In addition, the shared-panel LaCT NTP checkpoint was trained in a newer execution environment than the retained QA checkpoint. Appendix~\ref{app:protocol} records these distinctions.

\subsection{External evaluation}
\paragraph{BABILong.}
BABILong embeds bAbI reasoning tasks in longer distractor contexts \citep{babi,babilong}. We evaluate QA1--QA5, covering one-, two-, and three-supporting-fact questions, two-argument relations, and three-argument relations. For LaCT and DeltaNet, the panel uses the official 100-example release at nominal lengths 0K, 1K, 2K, 4K, and 8K: 2,500 examples per checkpoint. Official instructions, demonstrations, and post-prompts are enabled. Greedy generation is capped at 20 new tokens, and scoring uses the official answer-matching function. Nominal 0K denotes the shortest benchmark condition, not an empty context.

\paragraph{RULER.}
RULER tests long-context capabilities through controlled retrieval and related tasks \citep{ruler}. The shared single-needle panel uses 75 examples per cell: three single-needle tasks at 1K, 2K, and 4K, and two at 8K, totaling 825 examples per checkpoint. We preserve the official answer prefix, use greedy generation with a 128-token cap, and apply the official scorer. The task mixture changes at 8K; a higher aggregate score there need not imply that increasing context improves retrieval. We also report a complementary 500-example DeltaNet panel spanning multi-value retrieval, variable tracking, word aggregation, and document QA.

\paragraph{RWKV-7 panel.}
The separately reported RWKV-7 comparison evaluates 500 BABILong examples (QA1--QA5, 50 per task at 8K and 16K) and 600 RULER examples (three single-needle tasks, 100 per task at those lengths). Greedy generation permits 32 and 128 tokens, respectively. Actual native input lengths plus generation allowance fit within 16,384 tokens without truncation. These scores are not pooled with the shared 0--8K panel.

The evaluation uses official benchmark prompts and scoring with model-specific inference adapters. This preserves benchmark definitions without claiming that the upstream benchmark repositories natively implement every backbone. All comparisons use frozen examples within a panel. External answers never provide training labels.

\section{Results}
\subsection{Transfer beyond event prediction}
Table~\ref{tab:main} summarizes the three conditions. LaCT delayed QA improves over event NTP from 25.60\% to 31.08\% on BABILong and from 87.39\% to 88.85\% on single-needle RULER. The corresponding unrounded gains are 5.48 and 1.45 percentage points. Relative to the original checkpoint, the gains are 9.60 and 3.64 points. These improvements show that answer supervision can transfer from simulator-generated state questions to separately constructed long-context evaluations.

\begin{table}[t]
\centering
\caption{Official benchmark scores (\%). Bold denotes the delayed-QA method, not a significance claim. LaCT and DeltaNet share the 2,500-example BABILong and 825-example RULER panels. RWKV-7 uses separate 500/600-example panels and is author-reported; its original-model result is unavailable. Cross-backbone scores should not be read as a controlled ranking.}
\label{tab:main}
\begin{tabular}{llrr}\toprule
Backbone & Training & BABILong & RULER single-needle\\\midrule
\input{data/main_table.tex}\end{tabular}
\end{table}

DeltaNet exhibits a smaller but positive aggregate change: BABILong rises from 21.16\% to 22.48\%, and RULER from 80.24\% to 83.52\%. Its paired example-bootstrap interval is $[-0.16,2.84]$ points for BABILong and $[1.70,4.97]$ for RULER. Thus the BABILong point estimate is positive without excluding zero under this uncertainty measure. These intervals capture evaluation-example variation, not variation across training seeds.

The RWKV-7 corrective report records BABILong increasing from 22.60\% to 27.20\% and RULER from 68.83\% to 75.83\%. These gains extend the observed direction to a generalized delta-rule test-time learner, but the different training budget and evaluation panel preclude direct magnitude comparisons. The underlying RWKV predictions were not available in the present repository checkout; we distinguish this report-level evidence from the locally inspected LaCT and DeltaNet score records.

\begin{figure}[t]
\centering\includegraphics[width=\linewidth]{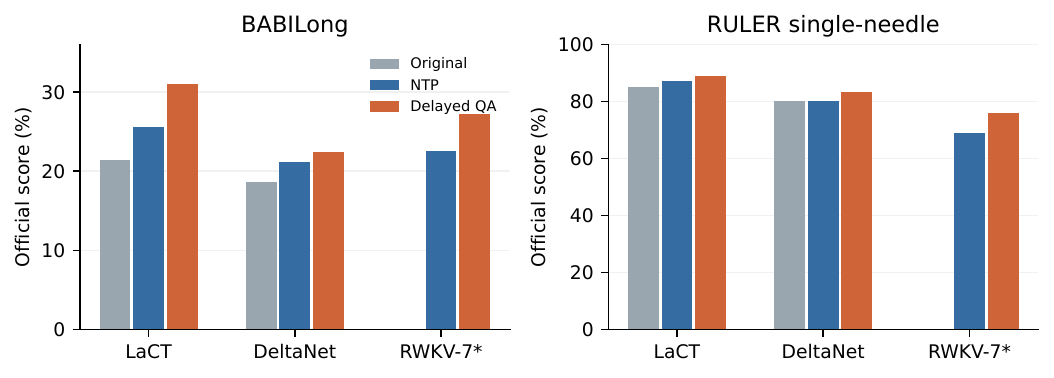}
\caption{Original, event-NTP, and delayed-QA scores. The gain over NTP is positive on both reported benchmarks for each backbone. RWKV-7 (*) is a separately reported comparison on different length panels; its original-model result is unavailable.}
\label{fig:overview}
\end{figure}

\subsection{Behavior as context grows}
\begin{figure}[t]
\centering\includegraphics[width=\linewidth]{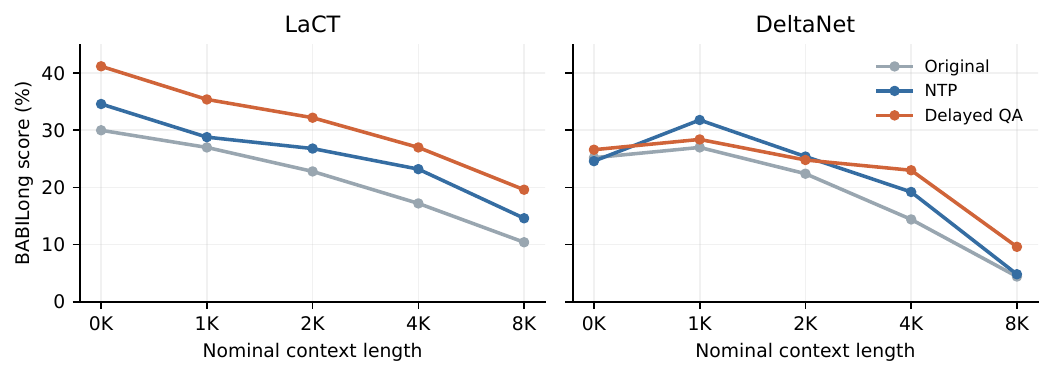}
\caption{BABILong by context length on the shared panel, with 500 examples per point. LaCT improves consistently over event NTP; DeltaNet's gains are concentrated at the longer evaluated lengths. Curves show one trained checkpoint per condition, not averages over seeds.}
\label{fig:length}
\end{figure}

Figure~\ref{fig:length} separates BABILong scores by nominal length. LaCT delayed QA exceeds NTP at every length in the shared panel, including 8K (19.60\% versus 14.60\%). For DeltaNet, the largest absolute gains occur at 4K and 8K: 23.00\% versus 19.20\%, and 9.60\% versus 4.80\%, respectively. DeltaNet nevertheless regresses at 1K and 2K, so its aggregate improvement is not uniform.

All conditions remain substantially less accurate on longer BABILong inputs. The auxiliary objective improves the evaluated operating range without eliminating the underlying long-context difficulty. In particular, the small absolute DeltaNet score at 8K should not be obscured by its large relative gain.

\subsection{Beyond single-needle retrieval}
The complementary DeltaNet panel tests whether improvement extends beyond locating a single target. Table~\ref{tab:broad} reports each task without collapsing heterogeneous metrics into an overall claim. Delayed QA improves variable tracking at both lengths, frequent-word extraction, and the two document-QA tasks. It decreases multi-value retrieval and common-word extraction. The result suggests selective transfer rather than a general increase in every memory capability.

\begin{table}[t]
\centering\small
\caption{Complementary DeltaNet RULER panel: 50 examples per cell. Scores are official task metrics, including partial-target credit for synthetic tasks; they are not whole-response exact match. The HotpotQA-based task belongs to RULER, not LongBench.}
\label{tab:broad}
\begin{tabular}{llrrr}\toprule Task & Length & Original & NTP & Delayed QA\\\midrule
Multi-value retrieval & 2K & 36.00 & 40.00 & 37.00\\
Variable tracking & 2K & 12.80 & 6.80 & 13.20\\
Common-word extraction & 2K & 19.20 & 13.40 & 12.80\\
Frequent-word extraction & 2K & 25.33 & 21.33 & 39.33\\
Multi-value retrieval & 4K & 18.50 & 20.00 & 14.50\\
Variable tracking & 4K & 18.00 & 8.00 & 22.40\\
Common-word extraction & 4K & 19.80 & 18.20 & 14.60\\
Frequent-word extraction & 4K & 12.00 & 16.00 & 27.33\\
SQuAD-based QA & 8K & 6.00 & 4.00 & 16.00\\
HotpotQA-based QA & 8K & 8.00 & 8.00 & 18.00\\\bottomrule
\end{tabular}
\end{table}

\section{Related work}
\paragraph{Learning at test time.}
Linear attention establishes recurrent computation with linear sequence-length scaling \citep{linear}. Fast-weight programming relates that computation to learned associative storage and motivates error-correcting updates \citep{fastweights}. DeltaNet develops efficient sequence-parallel training for the delta rule \citep{delta}; TTT layers, LaCT, and Titans expand the space of learned memory and adaptation mechanisms \citep{ttt,lact,titans}. RWKV-7 studies expressive recurrent state evolution \citep{rwkv}. Our method complements these architectural advances by changing the outer supervision that trains the test-time memory, without replacing its native update.

\paragraph{Supervision beyond the next token.}
ReFINE is particularly close in motivation: it augments NTP with reinforcement learning over next-sequence prediction and studies LaCT and DeltaNet \citep{refine}. Our auxiliary task instead uses simulator-verified semantic questions placed after a controlled delay. It requires no sampled reward optimization and directly supervises answer tokens. The contribution is not the general observation that objectives beyond NTP can help; it is a concrete state-grounded construction with retention and revision validity, causal branch isolation, and transfer evaluation.

\paragraph{Controlled environments and long-context evaluation.}
TextWorld supplies generated interactive environments with explicit state \citep{textworld}. The bAbI tasks isolate forms of question answering \citep{babi}, and BABILong embeds such reasoning in long distractor contexts \citep{babilong}. RULER complements this with controlled retrieval and additional long-context tasks \citep{ruler}. These resources allow a distinction between in-environment supervision and external evaluation, although both training and evaluation can share semantic skills such as tracking locations and resolving relations.

\section{Discussion and limitations}
The positive results support \emph{future semantic usefulness} as an outer training target for test-time memory. A delayed answer must be produced through the model's ordinary readout, which couples storage to the behavior ultimately evaluated. The method also accommodates revision: it rewards retaining the latest valid state rather than preserving every historical association indefinitely.

Several boundaries matter. First, the study compares the combined delayed-QA objective with original and event-NTP baselines. This comparison does not isolate the causal effect of delay. Existing timing-control experiments have mixed outcomes, so we do not claim that delayed questions universally outperform otherwise equivalent immediate questions. Second, answer-plus-EOS supervision also teaches formatting and stopping. Official benchmark improvements are meaningful under those metrics but are not, by themselves, proof of a purely mnemonic mechanism. Third, the runs use one training seed and previously exposed benchmark subsets. No claim of independent replication or full-suite performance follows from these panels.

Training is matched in event exposures within the principal comparisons, not in total computation: auxiliary branches require additional prefix processing. Cross-backbone recipes differ, and the retained LaCT QA reference and new NTP control differ in execution environment. The reported RWKV comparison still requires its prediction artifacts to be integrated into this repository for independent reproduction. Finally, TextWorld's symbolic structure makes labels reliable but narrows the training distribution. Transfer to unstructured factual revision, dialogue, and naturally occurring long histories remains an open question.

\section{Conclusion}
Delayed supervision trains test-time language models to preserve information for later semantic use. State questions provide this training signal after many intervening updates to the adaptive memory. The proposed objective preserves event NTP, supervises only answer tokens on isolated branches, and distinguishes retained facts from revised values. Across the reported LaCT, plain DeltaNet, and RWKV-7 comparisons, it improves aggregate BABILong and single-needle RULER scores over event-only training. The results motivate delayed semantic supervision as a simple complement to native test-time learning mechanisms, with further controlled replication needed to establish its generality and isolate the role of delay.

\section*{Reproducibility statement}
The accompanying repository contains training and evaluation code, pinned protocols, and the score records used for the LaCT and DeltaNet tables. The paper directory includes machine-readable result provenance and scripts that regenerate its figures. Appendix~\ref{app:protocol} specifies the comparison boundaries and model-specific training settings. RWKV-7 values are transcribed from the author-supplied corrective report and are explicitly identified as such.

\bibliography{references}
\bibliographystyle{iclr2026_conference}

\appendix
\section{Training and evaluation details}
\label{app:protocol}
\begin{table}[h]
\centering\small
\caption{Training settings for the reported QA arms. NTP controls use the same event exposure within each comparison and omit the auxiliary loss.}
\begin{tabular}{lrrr}\toprule
Setting & LaCT & DeltaNet & RWKV-7\\\midrule
Parameters & 760M & 1.3B & 1.5B\\
World exposures & 1,024 & 1,024 & 64\\
Effective batch (worlds) & 8 & 8 & 4\\
Outer updates & 128 & 128 & 16\\
QA branches / world & 12 & 8 & 2\\
Event coefficient $\alpha$ & 0.5 & 1 & 1\\
QA coefficient $\beta$ & 0.5 & 0.0158474 & 0.01\\
Peak learning rate & $10^{-5}$ & $10^{-5}$ & $10^{-5}$\\
AdamW betas & $(0.9,0.95)$ & $(0.9,0.95)$ & $(0.9,0.95)$\\
Weight decay & 0.01 & 0.01 & 0.01\\
Global gradient clip & 1 & 1 & 1\\\bottomrule
\end{tabular}
\end{table}

\paragraph{Initialization and native computation.}
LaCT uses the public 760M large-chunk model with 2,048-token native updates and local attention. DeltaNet uses \texttt{fla-hub/delta\_net-1.3B-100B}, pinned at revision \texttt{b4dcbbafd4fde802717bdec3008d4aba9cb3a1f8}, with the native FLA architecture and recurrence. The reported RWKV checkpoint is \texttt{RWKV/RWKV7-G1j-1.5B-20260831}, revision \texttt{2c18b29ab7fbece25ff6112281eea0fa41fcb30f}. Training uses FP32 weights with BF16 autocast in the documented recipes. No architecture is retrained from random initialization.

\paragraph{Auxiliary scaling.}
DeltaNet's coefficient was fixed before training as 0.25 divided by the largest measured unweighted QA/event gradient-norm ratio over the calibration conditions. The observed ratios ranged from 12.45 to 15.78. The resulting coefficient bounds the measured initial auxiliary norm contribution, but neither assigns a fixed fraction of the AdamW update nor establishes a universal optimal weight. RWKV's reported coefficient likewise reflects training-only gradient diagnostics. LaCT retains its original equal scalar allocation. These recipes must not be described as a common cross-model weighting ablation.

\paragraph{Data and validity.}
The DeltaNet paired-probe set has 4,096 unique questions: 2,457 location, 820 composition, 479 attribute, 171 counting, and 169 inventory questions. Retention versus revision is a separate classification. The symbolic answer is checked against an independent raw-fact oracle throughout the relevant interval. The terminal LaCT probes are additional end-of-stream questions and should not be conflated with the paired delayed-probe budget. No branch silently truncates its evidence to satisfy an input limit.

\paragraph{Benchmark scope.}
The LaCT and DeltaNet BABILong panels share the 25 frozen task--length cells. Their single-needle RULER panel uses the same frozen 825 inputs, prompts, answer prefixes, and generation allowances. Native token counts still differ by tokenizer; nominal benchmark lengths do not imply identical native sequence lengths. The complementary RULER panel uses 50 examples in each of ten cells. It is not the complete RULER suite, and its document-QA tasks are distinct from similarly named LongBench evaluations.

\section{Separately reported RWKV-7 length results}
The values below belong to the 16-update corrective comparison supplied by the authors. They do not come from the separate 64-update, $3\times10^{-6}$ run described elsewhere in the project history. We do not combine their checkpoints or scores.
\begin{table}[h]\centering
\begin{tabular}{llrr}\toprule Benchmark & Length & NTP & Delayed QA\\\midrule
BABILong & 8K & 24.00 & 29.20\\
BABILong & 16K & 21.20 & 25.20\\
RULER single-needle & 8K & 84.33 & 93.33\\
RULER single-needle & 16K & 53.33 & 58.33\\\bottomrule
\end{tabular}
\caption{RWKV-7 official scores (\%) from the supplied corrective report. Each BABILong row contains 250 examples and each RULER row 300.}
\end{table}

\section{Evidence provenance and research-assistance disclosure}
LaCT scores are taken from the complete shared-panel audit, and DeltaNet scores from the fixed final-update corrective comparison and complementary RULER audit. Figure data retain their source hashes. These checks verify consistency with saved result records; they are not a new execution of every benchmark prediction. RWKV-7 numbers are report-level evidence pending inclusion of the referenced prediction and verification files.

An AI coding and writing assistant helped inspect experiment records, retrieve bibliographic metadata, draft this manuscript, and produce its figures and build scripts. The assistant also participated in project implementation and diagnostic iteration. The authors are responsible for reviewing the experimental claims, citations, and final publication materials.

\section*{Copyright and disclosure notice}
\noindent\textcopyright\ 2026 Jinha Kim. All rights reserved.

Upon public posting on arXiv, this manuscript constitutes a public disclosure of the methods, experimental designs, and results described herein as of that posting date. Subsequent work that uses, extends, reproduces, or builds upon these contributions should appropriately cite this manuscript.

The text, figures, tables, and other original expressive content may not be reproduced or redistributed except as permitted by applicable law, under any license granted by the copyright holder (including the license granted to arXiv), or with prior permission from the copyright holder. This notice does not restrict arXiv's rights under the license selected for the submission.

\end{document}

%% file: data/main_table.tex
LaCT & Original & 21.48 & 85.21 \\
 & NTP & 25.60 & 87.39 \\
 & Delayed QA & \textbf{31.08} & \textbf{88.85} \\
\midrule
DeltaNet & Original & 18.68 & 80.36 \\
 & NTP & 21.16 & 80.24 \\
 & Delayed QA & \textbf{22.48} & \textbf{83.52} \\
\midrule
RWKV-7 & NTP & 22.60 & 68.83 \\
 & Delayed QA & \textbf{27.20} & \textbf{75.83} \\
\bottomrule